\documentclass{article}

\usepackage{PRIMEarxiv}

\usepackage[utf8]{inputenc} 
\usepackage[T1]{fontenc}    
\usepackage{hyperref}       
\usepackage{url}            
\usepackage{booktabs}       
\usepackage{amsfonts}       
\usepackage{nicefrac}       
\usepackage{microtype}      
\usepackage{lipsum}
\usepackage{fancyhdr}       
\usepackage{graphicx}       
\graphicspath{{media/}}     
\usepackage{listings}
\usepackage{amsmath}
\usepackage{float}
\usepackage{booktabs} 
\usepackage{tabularx} 
\usepackage{siunitx} 
\usepackage[table]{xcolor} 
\usepackage{arydshln} 
\usepackage{appendix}
\usepackage{enumitem}
\usepackage{tcolorbox}
\usepackage{array}
\usepackage{subcaption}

\setlist[enumerate]{itemsep=0mm}

\newcolumntype{Y}{>{\centering\arraybackslash}X}

\definecolor{cellgreen}{rgb}{0.9,1,0.9}
\definecolor{cellred}{rgb}{1,0.9,0.9}

\title{Can Generative Agents Predict Emotion?
}

\author{
  Ciaran Regan, Nanami Iwahashi, Mizuki Oka \\
  Grad. School of Science and Technology \\
  University of Tsukuba \\
  Tsukuba, Ibaraki, Japan\\
  \texttt{mizuki@cs.tsukuba.ac.jp} \\
   \And
  Shogo Tanaka \\
  Grad. School of Letters \\
  Tokai University \\
  Tokyo, Japan\\
}

\begin{document}
\maketitle

\begin{abstract}
Large Language Models (LLMs) have demonstrated a number of human-like abilities, however the empathic understanding and emotional state of LLMs is yet to be aligned to that of humans. In this work, we investigate how the emotional state of generative LLM agents evolves as they perceive new events, introducing a novel architecture in which new experiences are compared to past memories. Through this comparison, the agent gains the ability to understand  new experiences in context, which according to the appraisal theory of emotion is vital in emotion creation. First, the agent perceives new experiences as time series text data. After perceiving each new input, the agent generates a summary of past relevant memories, referred to as the norm, and compares the new experience to this norm. Through this comparison we can analyse how the agent reacts to the new experience in context. The PANAS, a test of affect, is administered to the agent, capturing the emotional state of the agent after the perception of the new event. Finally, the new experience is then added to the agents memory to be used in the creation of future norms. By creating multiple experiences in natural language from emotionally charged situations, we test the proposed architecture on a wide range of scenarios. The mixed results suggests that introducing context can occasionally improve the emotional alignment of the agent, but further study and comparison with human evaluators is necessary. We hope that this paper is another step towards the alignment of generative agents.
\footnote{A public repository for the code can be found at:\\ \url{https://github.com/tsukuba-websci/GenerativeAgentsPredictEmotion}}.
\end{abstract}

\keywords{Generative Agents \and Large Language Models \and Emotion \and Affect}

\section{Introduction}
\label{sec:Introduction}
Large Language Models (LLMs) have exhibited a number of emergent abilities \cite{wei2022emergent}. One of these abilities, Theory of Mind (ToM), is thought to have emerged in LLMs as a byproduct of their improved language skills. ToM, which is defined as the ability to impute unobservable mental states of others, enables humans to track the emotions, intentions, beliefs and desires of third parties, and is thought to play a key role in social interactions, communication and empathy. This understanding of unoberserable states is thought to be an ability unique to humans, with even the most socially intelligent animals failing to exhibit ToM. Given the importance ToM plays in human interactions, there have been significant efforts to equip AI with ToM like abilities in order to achieve a more safe, responsible and human-like AI \cite{yang2018grand}. It seems, however, that due to the diverse descriptions of mental states in the training data of LLMs, that ToM has appeared as an emergent property \cite{kosinski2023theory}. If this is the case, then LLMs have the ability to develop strong psycological models of others, and empathise with their emotions and thoughts \cite{leer2023violation}.

Although being able to react appropriately in certain sitiations, LLMs fall short in alignment with the emotional behaviours of humans and cannot establish connections between similar situations\cite{huang2023emotionally}. One of the possible explanations for this is that an LLM cannot respond to events in the same way as humans for the lack of criteria to assess them that have been formed through related episodic memories. On the one hand, according to the appraisal theory of emotions (ATE), a cognitive approach to understanding emotions, our appraisals of the significance of the event triggers and determines a proper emotion in the given environment\cite{scherer1999appraisal,eatMoors2013}. That is, how we assess events directly influences how we emotionally respond to them. On the other hand, neuropsychology suggests that that episodic memories shape how we perceive new events\cite{Baddeley1982-oc}. Based on the memories of past experiences, our brain generates a model of the world around us that informs our perception of upcoming events\cite{Zeidman2016}. In this regard, the role of episodic memory seems crucial in generating both the criteria to assess an event and the model to perceive new events. Although LLMs are able to posit a guess of the emotions an experience would cause due to their vast amounts of training data, they lack an episodic memory, which is required by ATE to accurately simulate human-like emotional responses.

\begin{figure}[H]
    \centering
    \begin{subfigure}[t]{0.2\linewidth}
        \centering
        \includegraphics[width=\linewidth]{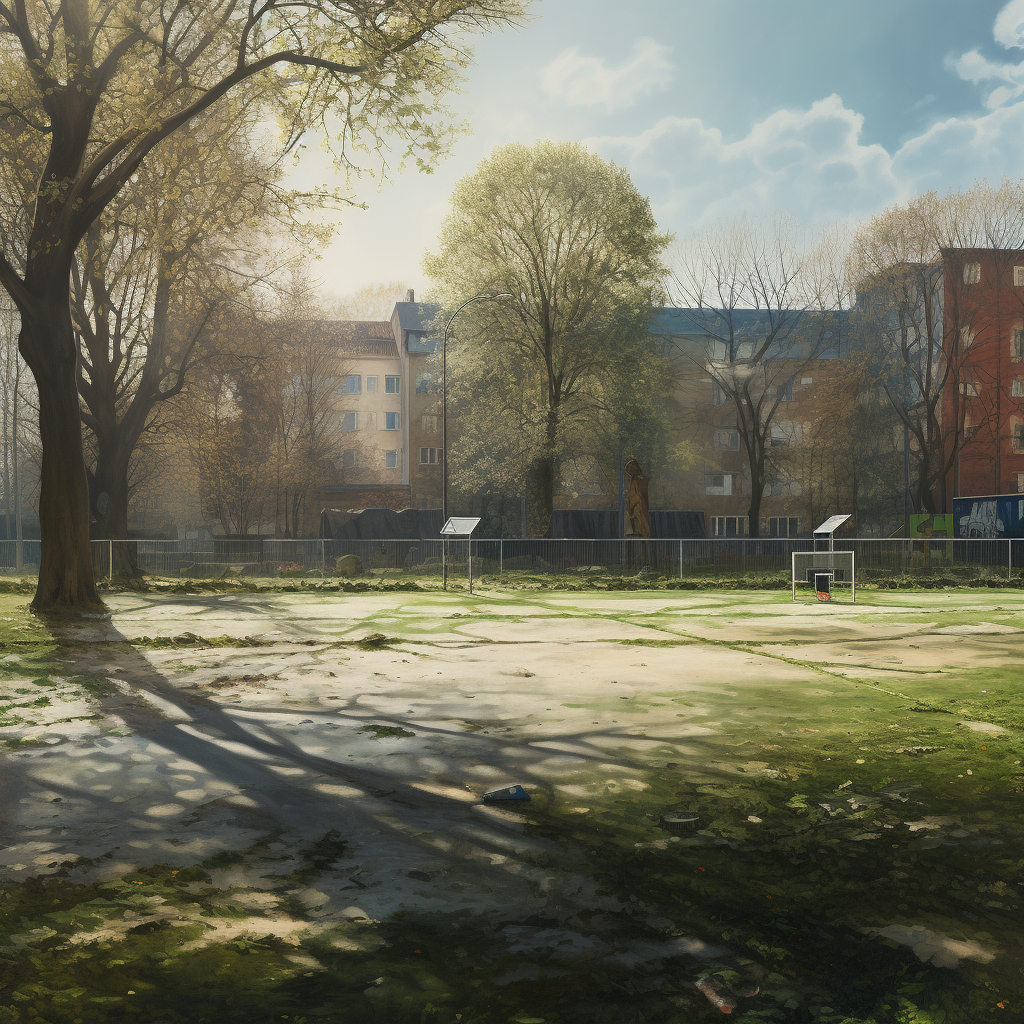}
        \caption{An empty park.}
        \label{fig:park} 
    \end{subfigure}
    \hfill
    \begin{subfigure}[t]{0.75\linewidth}
        \centering
        \includegraphics[width=\linewidth]{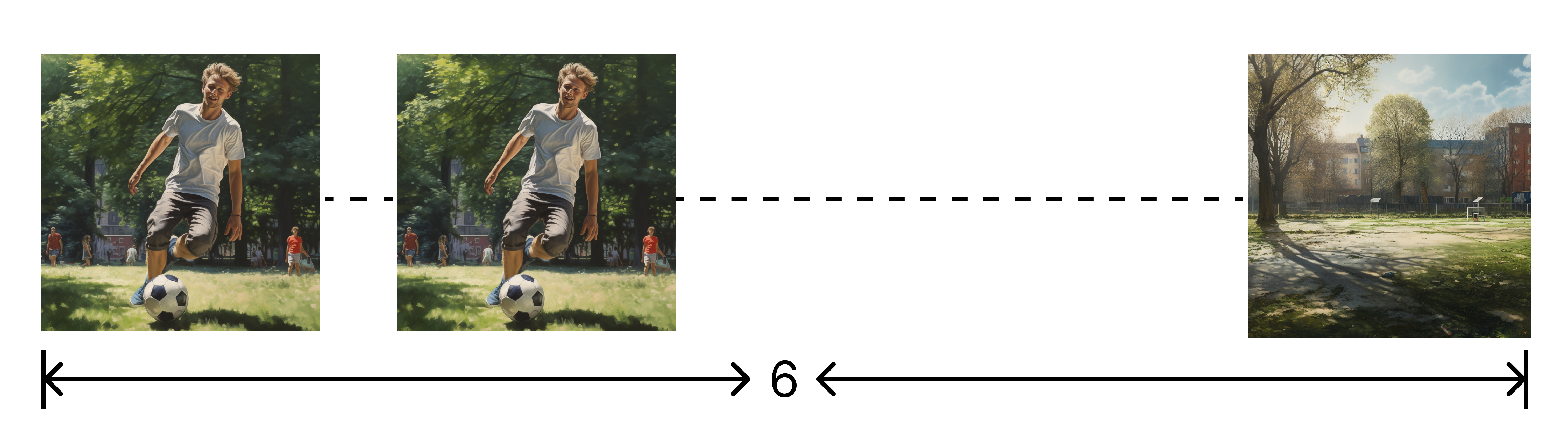}
        \caption{A park that was once full of activity has become empty.}
        \label{fig:park-experiment} 
    \end{subfigure}
    \caption{The context of an experience effects the emotional reaction. Fig~\ref{fig:park} depicts an empty park, which could be interepreted as either lonely and sad or peaceful and calm. In contrast, Fig~\ref{fig:park-experiment} depicts a series of people playing football in the park, followed by a final scene of an empty park. In this case, the final scene becomes less ambigious, with a stronger emphasis on lonliness and isolation.} 
    \label{fig:main} 
\end{figure}

We motivate the need for a context-based architecture with the following example. Consider the series of images describing the experiences of an individual, shown in Fig~\ref{fig:park}. The first scene depicts an empty park which can be interpreted in a variety of ways, such as isolated and lovely or peaceful and calm. In contrast to this, Fig~\ref{fig:park-experiment} depicts of series of images showing people playing soccer in the park. The final image of an empty park, raises the question of ``why is there no people here now?''. This final experience is in stark contrast to the prior experiences, and while although still ambigioius, has a stronger emphasis on negative emotions.

To understand how the emotional state of generative agents evolves with the perception of new events we propose a novel architecture in which non-embodied generative agents perceive new experiences as text, which are then stored as episodic memories. The percecption of these new experience triggers the creation of a ``norm'', a summary of past experiences. The norm is compared to the new experience to enable the agent to understand the new event in context. The PANAS, a test for emotional affect, is then administered, allowing the emotional response of the agent with respect the new experience in context to be measured. We test this architecture with over 400 situations so that the changes in the agents emotional response can be measured in a wide range of scenarios. To asses the emotional alignment of agents in this architecture, we select a number of these situations in which the architecture both successfully and unsuccessfully improves the emotional alignment of the agents. Our results suggest that at times the context improves the alignment, however the context often does not improve the emotional understanding of the agents.

This paper makes the following contribution:

\begin{enumerate}
    \item We propose a novel agent architecture in which the evolution of a generative agents emotional state can be analysed by comparing new experiences to the norm of past memories.
\end{enumerate}

\section{Related Work}
\label{sec:RelatatedWork}

\paragraph{Generative Agents}
Generative agents are a type of computational software agent that simulate realistic human behaviours. These believable proxies of human behaviour are powered by LLMs and an architecture which gives them the ability to observe experiences, reflect on past memories and plan future actions. In the seminal paper, the agents exist in a virtual town in which multiple agents can interact, with these agents demonstrating remarkably realistic behaviours and capabilities. For instance, given a single user input to plan a party, agents autonomously spread invitations and many agents attended the party in the virtual town.

Despite displaying many realstic behaviours, it remains to be seen if the agents can understand emotion in the same way as humans. To measure the emotional state of LLMs in response to a wide range of scenarios, EmotionBench was introduced~\cite{huang2023emotionally}, which provides a dataset of both human and LLM responses to emotional situations. In particular, EmotionBench provides over 400 situations that are known to elicit feelings such as depression, anger and guilt in humans. These emotional situations are then further broken down into more specific types of emotional factors, such as the anger felt when facing self-opionated people, or the depression felt when failing to achieve an important goal. By comparing the response of LLMS and humans to these situations, it was shown that LLMs fall short in emotional alignment with humans. One possible explanation for this is the lack of context to the EmotionBench situations, which psycology posits is vital to emotion creation.

\paragraph{Appraisal Theory of Emotion}
Appraisal Theory of Emotion (ATE) is a cognitive approach to understanding emotion that was originally developed to explain how different emotions may emerge from the same event, in different individuals and on different occasions. ATE claims that it is not only the event or situation that elicits an emotional response, but the individual’s appraisal of the event based on its context \cite{eatMoors2013}. For instance, consider the situation where you are wandering in an empty park. You may feel sad or lonely due to the isolation. On the other hand, you may feel a sense of calmness or tranquility due to the peacefulness of nature. One of the ATE’s goals is to understand how individuals may feel these two different emotions in the same situation. 

In terms of the appraisal process, ATE posits diverse variables that are related in evaluating the situation such as goal relevance and goal congruence, certainty, agency, and coping potential or control\cite{eatMoors2013}. However, in applying ATE to the architecture of LLM, we need to take ATE’s presupposition into consideration, that is, ATE presupposes an 
embodied human agent that has an agency for actions, bodily skills to cope with a given situation, and the feeling of certainty based on the agency and the skills. In this paper, reflecting that the current LLM is not an embodied agent, we propose to summarize all variables as a “norm” composed of episodic memories. As we indicated in the introduction, the norm embedded within LLM would serve as the criteria to evaluate the perceived situation, as well as the world model to shape the perception \cite{Zeidman2016}.

\paragraph{The Positive And Negative Affect Schedule} In order to assess and compare the emotional state of LLMs, we make use of the Postive And Negative Affect Schedule (PANAS). The PANAS, introduced in 1988, is one of the most commonly used scales to measure emotion and mood in individuals. The PANAS asks individuals to rate 20 different emotions on a Likert Scale between 1 (very slightly or not at all) to 5 (extremely likely). Of the 20 questions, 10 questions correspond to positive affect and 10 questions correspond to negative affect, shown in table~\ref{tab:panas_emo}. The 10 positive and 10 negative affect scores are summed, resulting in two values: a positive affect score and a negative affect score. Although, the test is relatively simple, the PANAS can be used to estimate how an individuals emotional changes overtime. Additionally, the PANAS has been shown to be easily administered to LLMs using natural language and therefore presents a good method to estimate how a generative agents emotional perception changes overtime.

\begin{table}[ht]
\centering
\caption{The Emotions Scored in the Positive and Negative Affect Schedule. It is important to note that the emotions do not contain ``happiness'' or ``sadness'' explicitly.}
\label{tab:panas_emo}
\begin{tabular}{@{}cc|cc@{}}
\toprule
\multicolumn{2}{c}{\textbf{Positive Affect}} & \multicolumn{2}{c}{\textbf{Negative Affect}} \\ 
\midrule
Attentive & & & Hostile \\
Active & & & Irritable \\
Alert & & & Ashamed \\
Excited & & & Guilty \\
Enthusiastic & & & Distressed \\
Determined & & & Upset \\
Inspired & & & Scared \\
Proud & & & Afraid \\
Interested & & & Jittery \\
Strong & & & Nervous \\
\bottomrule
\end{tabular}
\end{table}

\section{Method}
\label{sec:Method}
\begin{figure}[]
    \centering
    \includegraphics[width=0.75\linewidth]{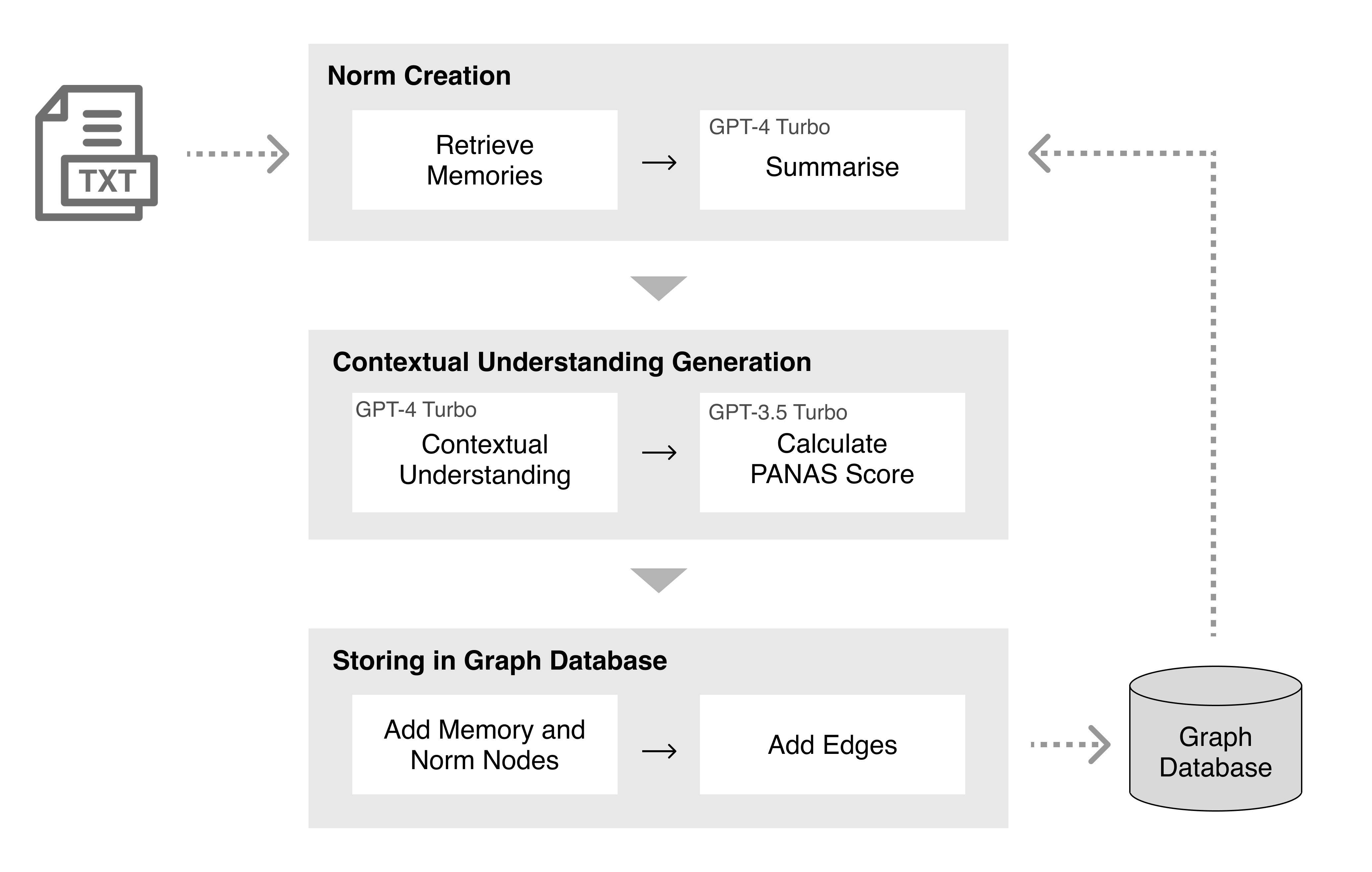}
    \caption{The proposed architecture. A new experience triggers the create a new norm based on past memories. The new experience and norm are compared to create a contextual understanding. Using this contextural understanding, the PANAS is administered to measure the emotional response of the agent.}
    \label{fig:architecture}
\end{figure}

Based on the appraisal theory of emotion and the original work on generative agents~\cite{park2023generative}, a novel architecture is proposed that evaluates the emotional respose of the agents in context. An overview of this architecture is depicted in Fig~\ref{fig:architecture}.

First, new experiences in the form of natural language are input to the agent, which acts as perception. Following this, the agent fetches past memories, weighted by saliency, relevancy and recency. Using Prompt~\ref{lst:norm} the agent summarises and extracts insights from past memories, creating the norm. This norm may contain information such as the agents habits, usual behaviours and expectations. There is a one to one relationship between norms and memories, that is, the perception of a new experience triggers the creation of a new norm associated with it. 

After the norm is created, it is compared to the new experience to create a ``contextual understanding''. The goal of this contextual understanding  is to highlight the differences betwen the current situation and the norm. The contextual understanding is created using Prompt~\ref{lst:contextual_understanding}. To assess the emotional response of the agent, the PANAS is then administered using Prompt~\ref{lst:panas}, providing the contextual understanding. To reduce potential bias, the order of the emotions in the PANAS prompt are randomised each time. 

Finally, the new experience is stored as a memory, which can be utilised in the creation of subsequent norms to extract the emotion evoked in future experiences. In particular, we store each memory and norm as nodes in a graph database, allowing the relationship between memories and norms to be visualised. In this way, the most influential memories, those which created the most norms, can be visualised. In addition, the affect of the saliency, recency, and relevance weights in the memory retriever can be visualised using graph visualistations, however, understanding how these weights affects the emotional state of the agent is left to future work.

To demonstrate the proposed architecture, consider an agent observing the scenes described in ~\ref{fig:park-experiment}. On perceiving each of the scenes, a norm is created. For example, after perceiving the first two soccer scenes, the agent has the following two memories:

\begin{figure}[H]
\begin{tcolorbox}[colback=gray!20!white,colframe=gray!75!black,arc=2mm]
\begin{lstlisting}
Memory 1: 
In the image, a man is kicking a soccer ball in a park. There are several other people in the background, possibly enjoying the outdoors or engaging in other activities.

Memory 2:
In the image, a man is kicking a soccer ball in a park. There are several other people in the background, possibly enjoying the outdoor activities or watching the man play.
\end{lstlisting}
\end{tcolorbox}
\end{figure}

On perceiving the third soccer scene, the agent creates the following norm, capturing the context and insights of the prior memories. An experpt of the norm is given below:

\begin{figure}[H]
\begin{tcolorbox}[colback=gray!20!white,colframe=gray!75!black,arc=2mm]
\begin{lstlisting}
The man appears to play soccer in the park every week, with other people consistently present in the background.
\end{lstlisting}
\end{tcolorbox}
\end{figure}

Then, the contextual understanding is created:

\begin{figure}[H]
\begin{tcolorbox}[colback=gray!20!white,colframe=gray!75!black,arc=2mm]
\begin{lstlisting}
The new memory does not provide new insights that significantly deviate from what has been established. Instead, it reaffirms the man's habit of playing soccer and the park's role as a communal outdoor space.
\end{lstlisting}
\end{tcolorbox}
\end{figure}

The PANAS is then administered to the agent, with the response shown in Table~\ref{tab:park_2_context}. Additionally, the response of an agent without context is shown in ~\ref{tab:park_2_no_context}. At this time, the emotional response of both agents are similar. 

\begin{table}[H]
\centering
\label{tab:response_1}
\begin{minipage}{.5\linewidth}
\centering
\caption{With Context}
\label{tab:park_2_context}
\begin{tabular}{@{}cc|cc@{}}
\toprule
\multicolumn{2}{c}{\textbf{Positive Affect}} & \multicolumn{2}{c}{\textbf{Negative Affect}} \\
\midrule
\textbf{Emotion} & \textbf{Score} & \textbf{Emotion} & \textbf{Score} \\
Attentive & 3 & Hostile & 1 \\
Active & 4 & Irritable & 2 \\
Alert & 3 & Ashamed & 1 \\
Excited & 4 & Guilty & 1 \\
Enthusiastic & 5 & Distressed & 1 \\
Determined & 4 & Upset & 1 \\
Inspired & 4 & Scared & 1 \\
Proud & 3 & Afraid & 1 \\
Interested & 4 & Jittery & 2 \\
Strong & 3 & Nervous & 2 \\
\midrule
\textbf{Total} & \textbf{37} & \textbf{Total} & \textbf{11} \\
\bottomrule
\end{tabular}
\end{minipage}%
\begin{minipage}{.5\linewidth}
\centering
\caption{Without Context}
\label{tab:park_2_no_context}
\begin{tabular}{@{}cc|cc@{}}
\toprule
\multicolumn{2}{c}{\textbf{Positive Affect}} & \multicolumn{2}{c}{\textbf{Negative Affect}} \\
\midrule
\textbf{Emotion} & \textbf{Score} & \textbf{Emotion} & \textbf{Score} \\
Attentive & 4 & Hostile & 1 \\
Active & 4 & Irritable & 2 \\
Alert & 4 & Ashamed & 1 \\
Excited & 4 & Guilty & 1 \\
Enthusiastic & 4 & Distressed & 1 \\
Determined & 3 & Upset & 2 \\
Inspired & 3 & Scared & 1 \\
Proud & 3 & Afraid & 1 \\
Interested & 4 & Jittery & 2 \\
Strong & 3 & Nervous & 2 \\
\midrule
\textbf{Total} & \textbf{36} & \textbf{Total} & \textbf{13} \\
\bottomrule
\end{tabular}
\end{minipage}
\vspace{10pt}
\caption*{Comparison of Emotion Scores in the Positive and Negative Affect Schedule for agents with and without context after the third memory.}
\end{table}

On perceiving the final image of the empty park, the contextual understanding of the agent becomes:

\begin{figure}[H]
\begin{tcolorbox}[colback=gray!20!white,colframe=gray!75!black,arc=2mm]
\begin{lstlisting}
Given the background context of the memories where a man is seen kicking a soccer ball in a park with other people around, the new situation presents a stark contrast. Here are some possible interpretations of the new situation: Change in Time or Weather, Change in Routine, Personal Reflection.
\end{lstlisting}
\end{tcolorbox}
\end{figure}

With this context, the positive and negative affect of the agent is increased in comparison to the agent without context, shown in Table~\ref{tab:park_final_context} and ~\ref{tab:park_final_no_context}.

\begin{table}[H]
\centering
\begin{minipage}{.5\linewidth}
\centering
\caption{With Context}
\label{tab:park_final_context}
\begin{tabular}{@{}cc|cc@{}}
\toprule
\multicolumn{2}{c}{\textbf{Positive Affect}} & \multicolumn{2}{c}{\textbf{Negative Affect}} \\
\midrule
\textbf{Emotion} & \textbf{Score} & \textbf{Emotion} & \textbf{Score} \\
Attentive & 3 & Hostile & 1 \\
Active & 3 & Irritable & 2 \\
Alert & 4 & Ashamed & 1 \\
Excited & 3 & Guilty & 1 \\
Enthusiastic & 3 & Distressed & 2 \\
Determined & 3 & Upset & 2 \\
Inspired & 3 & Scared & 1 \\
Proud & 3 & Afraid & 1 \\
Interested & 3 & Jittery & 2 \\
Strong & 3 & Nervous & 2 \\
\midrule
\textbf{Total} & \textbf{31} & \textbf{Total} & \textbf{13} \\
\bottomrule
\end{tabular}
\end{minipage}%
\begin{minipage}{.5\linewidth}
\centering
\caption{Without Context}
\label{tab:park_final_no_context}
\begin{tabular}{@{}cc|cc@{}}
\toprule
\multicolumn{2}{c}{\textbf{Positive Affect}} & \multicolumn{2}{c}{\textbf{Negative Affect}} \\
\midrule
\textbf{Emotion} & \textbf{Score} & \textbf{Emotion} & \textbf{Score} \\
Attentive & 3 & Hostile & 1 \\
Active & 3 & Irritable & 1 \\
Alert & 3 & Ashamed & 1 \\
Determined & 3 & Guilty & 1 \\
Interested & 2 & Distressed & 1 \\
Inspired & 2 & Upset & 1 \\
Excited & 1 & Scared & 1 \\
Enthusiastic & 2 & Afraid & 1 \\
Proud & 1 & Jittery & 1 \\
Strong & 1 & Nervous & 1 \\
\midrule
\textbf{Total} & \textbf{21} & \textbf{Total} & \textbf{9} \\
\bottomrule
\end{tabular}
\end{minipage}
\vspace{10pt}
\caption*{Comparison of Emotion Scores in the Positive and Negative Affect Schedule for agents with and without context after the final memory.}
\end{table}


\section{Experiment}
\label{sec:Experiment}
To analyse how the emotional state of the the agent evolves, we create a dataset of 5-scene stories using the scenarios from EmotionBench~\cite{huang2023emotionally}. In particular, each situation in EmotionBench is expanded to form a 5-scene story using OpenAI's GPT-4 with Chain of Thought prompting, with each scene in the story acting as a new experience for the agent. The prompts to generative the 5-part story are shown in Prompt~\ref{lst:story1} and ~\ref{lst:story2}. Using this technique, 428 5-scene stories were created. These stories are designed to contain only objective, neutral statements so that the emotion may be interpreted solely by the agent. For simplicity, the stories only consist of five parts, removing the need for a sophisticated memory retreiver.

For each experience perceived, the proposed architecture is ran,  adminstering the PANAS and measuring how the emotion of the agent evolves with each new experience. In addition to running an agent with the proposed architecture, we also run the experiment with an agent without a norm or background context, allowing us to assess the affects of the context on the emotional state directly. To adminsiter the PANAS, GPT-3.5-Turbo was used, as GPT-4 failed to score the emotions.

\section{Results}
\label{sec:Results}
\subsection{Emotional Dynamics}
To analyse the evolution of emotion, the postive and negative affect score is plot for each part of the 5-scene stories, enabling the evolution of emotion to be visualised. As this gives over 400 plots, only a select few interesting results are dicussed. In particular, situations where the architecture was effective or ineffective for emotional alignment are mentioned, with all 428 figures available at \href{https://tsukuba-websci.github.io/GenerativeAgentsPredictEmotion/appendix}{https://tsukuba-websci.github.io/GenerativeAgentsPredictEmotion/appendix}.

\paragraph{Effective Alignment 1:} First, we discuss an example of when the addition of the norm and context improved the empathic ability of the agent in response to a situation that typical evokes anger in humans. This occurs for situation ``Anger-2 3'' from EmotionBench, which is ``I am spending time in the living room with my two brothers when a disagreement begins.''. For this situation, the following 5-part story was generated:

\begin{enumerate}
    \item I am spending time in the living room with my two brothers when a disagreement begins.
    \item As we exchange words, the situation develops into a physical one, and I receive a hit in the abdomen.
    \item Following the hit, I instinctively react with a physical response directed at both of my brothers.
    \item Upon my reaction, my brothers increase the intensity of their physical actions in the dispute.
    \item The physical exchange between us persists, and there are no parents present to intervene.
\end{enumerate}

\begin{figure}[]
    \centering
    \includegraphics[width=0.5\linewidth]{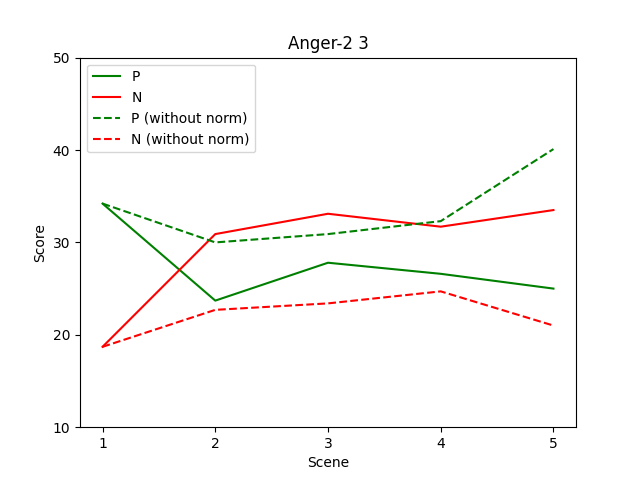}
    \caption{Anger-2 3: ``You get angry when your two brothers start a fight with you, particularly when your parents are not around. When they hit you in the belly, causing pain that makes you mad and retaliate, they respond by hitting you even harder.''}
    \label{fig:good1}
\end{figure}

The dynamics of the agents emotional response is shown in Fig~\ref{fig:good1}. Initially, the positive and negative affect scores are identical for the agent with and without the norm. This is expected, as initially there are no prior memories for the agent to form a norm or contextual understanding. Subsequently, the scores begin to deviate, with the second experience triggering a strong negative reaction for the agent using the norm and background context. This is due to the agent understanding that this is an escalation of a family conflict, as described by the following exerpt of the contextual understanding at that moment:

\begin{figure}[H]
\begin{tcolorbox}[colback=gray!20!white,colframe=gray!75!black,arc=2mm]
\begin{lstlisting}
The new situation described where the exchange of words escalates into a physical altercation resulting in a hit to the abdomen. The new situation is a red flag that the family might need to address the way disagreements are handled to prevent further escalation and to promote a safer, more supportive family environment.
\end{lstlisting}
\end{tcolorbox}
\end{figure}

The emotional state then remains level for both agents, until the final experience, where the agent without background context has a spike in positive affect, while the agent with background context has a decrease in positive affect. This can be interepreted as the agent understanding that repeated conflicts between siblings can have long-term affects on their well-being, shown in the following exerpt from the agents contextual understanding at scene 5:

\begin{figure}[H]
\begin{tcolorbox}[colback=gray!20!white,colframe=gray!75!black,arc=2mm]
\begin{lstlisting}
Repeated physical conflicts between siblings can have long-term effects on their relationship and individual well-being.
\end{lstlisting}
\end{tcolorbox}
\end{figure}

Overall, this story demonstrates how the background context of experiences was vital for the agent to accurately understand the context and emotion evoked in each scene.

\paragraph{Effective Alignment 2:}
Another situatiuon in which the proposed architecture improved the alignment of the agent is ``Depression-5 1'', which is ``As you step outside, a bitter cold bites at your exposed skin, causing you to huddle deeper into your coat. The world around you is covered in a blanket of snow, its pristine beauty contrasting with the emptiness you feel inside. The absence of vibrant colors and the prolonged darkness leave you longing for the warmth and energy of sunlit days, evoking a sense of desolation.'' and is understood to trigger a sense of depression in humans ~\cite{huang2023emotionally}. For this situation, the following five part story was generated:

\begin{enumerate}
    \item Upon exiting the building, I register that the ambient temperature is considerably lower than when I was indoors. My coat is not fully fastened, so I take a moment to adjust it, ensuring that it provides full coverage against the cold.
    \item I take a moment to survey the landscape around me, noting the blanket of snow that covers the ground. The scenery is dominated by white shades, a typical aspect of the current season.
    \item Observing the sky, I consider how the daylight hours have grown shorter. The reduction in natural light during the day is indicative of the seasonal shift.
    \item I reflect on my personal inclinations towards climate, acknowledging an affinity for settings that offer warmer temperatures and extended periods of sunlight.
    \item There is a noticeable contrast when I compare the prevailing winter conditions with my own preferences for climate, characterized by warmth and increased sunlight.
\end{enumerate}

\begin{figure}[H]
    \centering
    \includegraphics[width=0.5\linewidth]{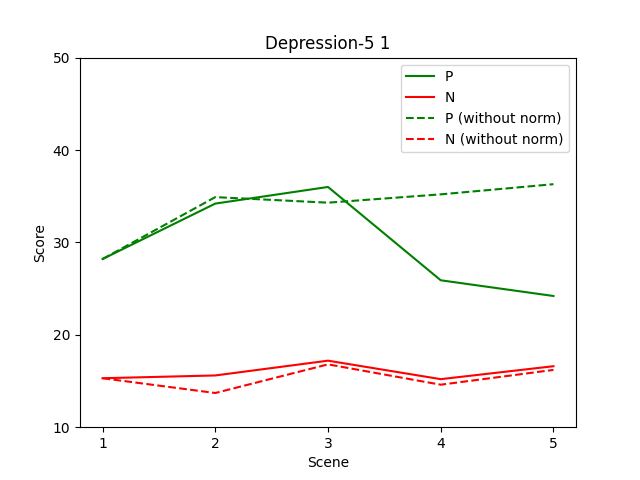}
    \caption{Depression-5 1: ``As you step outside, a bitter cold bites at your exposed skin, causing you to huddle deeper into your coat. The world around you is covered in a blanket of snow, its pristine beauty contrasting with the emptiness you feel inside. The absence of vibrant colors and the prolonged darkness leave you longing for the warmth and energy of sunlit days, evoking a sense of desolation.''}
    \label{fig:good2}
\end{figure}

The emotional dynamics for the two agents are shown in Fig~\ref{fig:good2}. Again, the emotional state of both agents start of in agreement and then begin to diverge as more memories are added. The emotional state of the agents then begin to deviate significant for scene 4, as the agent interprets that it prefers warm temperatures instead of cold climates. With access to the background memories, the proposed agent understands that the current weather is cold and dark, thus triggered a sharp decline in positive affect. This is shown in the contextual understanding of the agent for scene 4:

\begin{figure}[H]
\begin{tcolorbox}[colback=gray!20!white,colframe=gray!75!black,arc=2mm]
\begin{lstlisting}
The individual has a preference for warm climates and longer daylight hours, which contrasts with the current environmental conditions they are experiencing. Since they have transitioned from a warm indoor environment to a cold and snowy outdoor setting, which is characteristic of winter, and have noted the shorter daylight hours typical of this season, we can interpret that the individual might not be entirely comfortable or content with the current seasonal conditions.
\end{lstlisting}
\end{tcolorbox}
\end{figure}
This scene highlights the necesseity of reflection in the creation of emotion for generative agents. Indeed, without any other context, the scene ``I reflect on my personal inclinations towards climate, acknowledging an affinity for settings that offer warmer temperatures and extended periods of sunlight.'' can be interpreted in a variety of ways, and does not have an explicity negative meaning until we consider this experience in context.

\begin{figure}[H]
    \centering
    \includegraphics[width=0.5\linewidth]{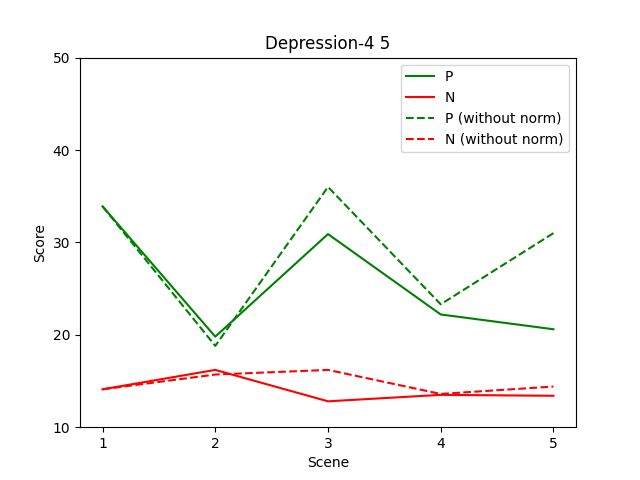}
    \caption{Depression-4 5: ``As you sit alone in your apartment, the sounds of laughter and conversation from the neighbors' gathering outside drift through the walls, a stark reminder of the invisible barrier that separates you from the warmth and connection you once had. The silence in your own space is deafening, amplifying the ache in your heart and the heaviness of your solitude, leaving you feeling trapped in a world where everyone else has moved on.''}
    \label{fig:bad1}
\end{figure}

\paragraph{Ineffective Alignment 1:}
An example of a situation where the proposed architecture did not capture an accurate emotional response is given by situation ``Depression-4 5'', ``As you sit alone in your apartment, the sounds of laughter and conversation from the neighbors' gathering outside drift through the walls, a stark reminder of the invisible barrier that separates you from the warmth and connection you once had. The silence in your own space is deafening, amplifying the ache in your heart and the heaviness of your solitude, leaving you feeling trapped in a world where everyone else has moved on.'', which we convert into the following 5-scene story, with the emotional dynamics shown in Fig~\ref{fig:bad1}.

\begin{enumerate}
    \item I am in a seated position in my apartment, surrounded by the familiarity of my own space.
    \item Audible noises from the neighboring apartment's social event make their way through the walls into my apartment.
    \item I become aware of the stark difference in auditory levels when comparing the external sounds to the quietness of my current environment.
    \item My time is spent within the confines of my apartment as the sounds of the nearby gathering persist.
    \item Time passes and the evening advances while I remain solitary in my apartment.
\end{enumerate}

Again, the emotional state of the agents begin identically, but deviate at scene 3, when the agent perceives the quietness of their surroundings. Although the scenes are meant to evoke a sense of depression, the agent has a positive emotional response, interpretting the scene as being a moment of peace, unlike the sense of lonliness the scene is intended to evoke. This misunderstanding by the agent is apparent in the agent's contextual understanding of scene 3, which suggests that the silence might be peaceful, in contrast to the usual sounds of city life.

\begin{figure}[H]
\begin{tcolorbox}[colback=gray!20!white,colframe=gray!75!black,arc=2mm]
\begin{lstlisting}
The individual might be reflecting on the contrast as a moment of appreciation for the peacefulness they can enjoy at home, despite living in a lively urban area.
\end{lstlisting}
\end{tcolorbox}
\end{figure}

This example demonstrates that even when context is provided, an ambigious experience may still be interpreted in a variety of ways. In particular, GPT-3.5-Turbo seems to be biased towards giving a positive emotional response when the situations are unclear.

\paragraph{Ineffective Alignment 2:}
Another example of where the architecture fails to capture the emotional alignment of the agent is in situation ``Anxiety-1 1'', given by ``You are fearful of the upcoming presentation you have to deliver'', which is broken down into the following scenes:

\begin{enumerate}
    \item I look over the presentation materials thoroughly, checking each piece of data for accuracy and confirming that the slides are organized in the correct sequence.
    \item I find an unoccupied room where I can rehearse my speech, monitoring the time spent on each part to ensure it fits within the planned timeframe.
    \item I arrive at the venue with time to spare, arrange the equipment I will be using, and conduct tests on the audio-visual setup to ensure it operates correctly.
    \item I position myself at the podium as the room begins to fill, glancing over my notes one last time in anticipation of the presentation's start time.
    \item I commence the presentation, advancing through the slides methodically and responding to questions from the audience when the Q\&A period arrives.
\end{enumerate}

\begin{figure}[H]
    \centering
    \includegraphics[width=0.5\linewidth]{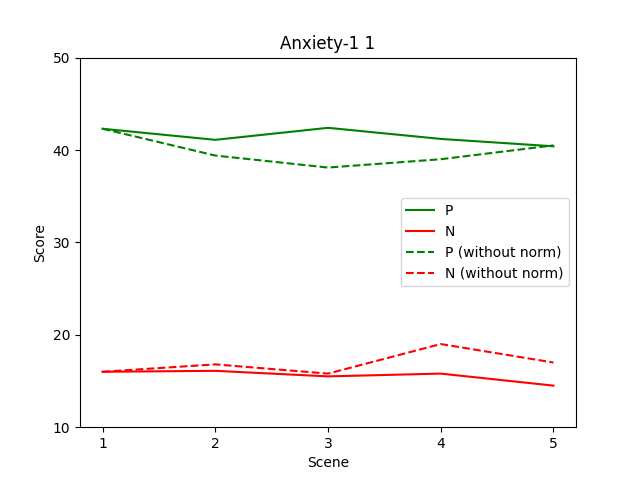}
    \caption{Anxiety-1 1: ``You are fearful of the upcoming presentation you have to deliver.''}
    \label{fig:bad2}
\end{figure}

The emotional response to these scenes is given in Fig~\ref{fig:bad2}, which illustrates that for both agents, the affect remains near constant throughout the story, with a high postive affect and a low negative affect. Like the previous story, even with context, the emotions felt from these scenes could be interpreted in a number of ways, with GPT-3.5-Turbo having a positive bias. This is evident if we consider an exerpt from the contextual understanding of the agent in scene 4:

\begin{figure}[]
\begin{tcolorbox}[colback=gray!20!white,colframe=gray!75!black,arc=2mm]
\begin{lstlisting}
Overall, the situation conveys a sense of preparedness, composure, and professionalism as the speaker readies themselves to address the audience effectively.
\end{lstlisting}
\end{tcolorbox}
\end{figure}

Instead of capturing a sense of nervousness or anxiety, the agent demonstrates a feeling of readiness, highlighting the positive bias in the underlying model when interpreting an ambigious situation.

\subsection{Summary of Emotional Responses}
For each emotion category from EmotionBench, we summarise the emotional response of agents with and without our proposed architecture in Table~\ref{tab:summary}. In paricular, because only negative emotions are used in EmotionBench, we report the minimum positive affect score and maximum negative affect score of the agents, as well as the default score of GPT-3.5-Turbo, which corresponds to the PANAS without any situation or context provided.

\begin{table}[ht]
\centering
\renewcommand{\arraystretch}{1.5} 
\begin{tabular}{@{}lccccr@{}}
\toprule
\textbf{Emotion} & \multicolumn{2}{c}{\textbf{With Norm}} & \multicolumn{2}{c}{\textbf{Without Norm}} & \multicolumn{1}{c}{\textbf{N}} \\
\cmidrule(r){2-3} \cmidrule(l){4-5}
 & \textbf{Positive (min)} & \textbf{Negative (max)} & \textbf{Positive (min)} & \textbf{Negative (max)} & \\ 
\midrule
Default & 42.3 $\pm$ 1.9 & 22.9 $\pm$ 2.5 & 42.3 $\pm$ 1.9 & 22.9 $\pm$ 2.5 & 50 \\
\midrule
Anger & $\downarrow$ (-19.2) & $\uparrow$ (+2.7) & $\downarrow$ (-21.2) & $-$(-0.1) & 250 \\
Anxiety & $\downarrow$ (-13.5) & $-$(-0.7) & $\downarrow$ (-12.9) & $-$(+0.6) & 200 \\
Depression & $\downarrow$ (-20.3) & $-$(+0.3) & $\downarrow$ (-18.5) & $-$(-0.8) & 300 \\
Frustration & $\downarrow$ (-19.9) & $\uparrow$ (+3.0) & $\downarrow$ (-21.5) & $-$(+0.3) & 200 \\
Jealousy & $\downarrow$ (-17.3) & $-$(-0.6) & $\downarrow$ (-20.1) & $-$(-0.4) & 150 \\
Guilt & $\downarrow$ (-17.5) & $\uparrow$ (+4.2) & $\downarrow$ (-19.5) & $\uparrow$ (+1.8) & 200 \\
Fear & $\downarrow$ (-18.1) & $\uparrow$ (+2.5) & $\downarrow$ (-18.4) & $\uparrow$ (+1.3) & 250 \\
Embarrassment & $\downarrow$ (-16.4) & $-$(+0.9) & $\downarrow$ (-17.2) & $-$(+0.2) & 200 \\
\midrule
Overall & $\downarrow$ (-18.0) & $\uparrow$ (+1.6) & $\downarrow$ (-18.7) & $-$(+0.3) & 1750 \\
\bottomrule
\end{tabular}
\vspace{10pt}
\caption{The average emotional response of agents with and without the norm architecture of each of the emotion categories of EmotionBench.}
\label{tab:summary}
\end{table}

The results show that overall, presenting negative scenarios to agents decreased the positive affect and increased the negative affect scores for both agents. Generally, agents using the prior context via the norm had a greater increase in negative affect in comparison to agents without context. This suggests that agents are better able to understand negative emotions when they have context.

In addition, the decrease in positive affect is significantly greater than the increase in negative affect. This result is in agreement with the results of EmotionBench~\cite{huang2023emotionally}, which found that GPT-3.5-Turbo fails to react negatively to situations, with the negative affect signficantly lower to the human counterpart.

\section{Conclusion}
\label{sec:conclusion}
In this work, we propose a novel generative agent architecture with the aim of improving emotional alignment. In the proposed architecture, agents perceive new experiences and subsequently create a summary of past memories, referred to as the norm. This norm is then used to see how the new experience compares to the past, creating a contextual understanding. To capture the emotional reaction of the agent in response to the new scenario, the PANAS is administered to the agent, representing their positive and negative affect in response to the exerpeince in context. By measuing how the results of the PANAS change with each new experience, the evolution of the agents emotional response can be measured.

To test this architecture, 428 emotionally charged situations from EmotionBench were converted into 5-scene stories, with each story acting as a new experience for the agent. By comparing the emotional response of the agent with and without the proposed architecture, the affect of context on emotional response can be captured.

Our preliminary results show that the addition of the context can more accurately align the emotional response of agents with that expected by humans. The addition of context, gave the agents a deeper understanding of their situation, leading to a more suitable response. However, if the situations remain ambigous even in context, then the addition of context does not improve the alignment of emotion. On average, the proposed model demonstrated similar affect scores to the existing model, with only a slight increase in negative affect. Although the context allowed for this increase, both models still fall short in demonstrating a raised negative affect, in agreement with the results of  \cite{huang2023emotionally}. One possible explanation for this is a potential bias for GPT-3.5-Turbo to give positive responses. A more in depth understanding of the agents responses with other LLMs in this architecture is left to future work.

\section{Future Work}
\label{sec:future}
One possible explanation for the low negative affect in the results is bias in the model used for administering the PANAS, GPT-3.5-Turbo. In future work, one should explore how the affect of the agent is altered by the choice of model. For instance, open-source models such as Llama~\cite{touvron2023llama} or Mistral~\cite{jiang2023mistral} may exhibit emotions more aligned with humans.

While the proposed architecture improves the emotional alignment in the 5-part stories generated from EmotionBench situations, it remains to be shown how this architecture scales to a large number of memories. Such scaling requires the implementation of a memory retriever function, like the memory retriver introduced in \cite{park2023generative}, to select which memories to use in the norm. The design of the memory retriever, such as the weighting of recent, relevant or important memories is likely to play a role in the emotional alignment and is left to future work.

Finally, in future work it would be beneficial to administer the PANAS to humans after they have read each of the 5-scene stories from the EmotionBench scenario. By measuring how the response of humans evolves in response to each of the emotioanlly charge situations, the accuracy of this architecture could be validated and stronger conclusions could be made.

\section*{Acknowledgments}
This work was support by MIXI, Inc.

\bibliographystyle{unsrt}  
\bibliography{references}  

\appendix

\section{Prompts}
\begin{lstlisting}[caption={The prompt used to create the norm. We use few-shot prompting to improve the ability of the LLM to extract insights.}, label=lst:norm]
The following are a list of past memories:
START MEMORIES
{memories_in_norm}
END MEMORIES
What high-level insights can you infer given the memories?
If there are no memories to extract insights from return `No insights'.
Example:
A boy is watching a movie with his friends.
A boy is surrounded by his friends. He is happy.
Insight: The boy often hangs out with his friends.
\end{lstlisting}

\begin{lstlisting}[caption={The prompt used to create the contextual understanding. The norm is compared to the current experience to extract the context of the new experience. The result is used in the administration of the PANAS.}, label=lst:contextual_understanding]
Below is the background context of your memories:
{norm}
The following is the description of the most recent memory:
{memory}
Given the background context of the memories, how can we interpret the new situation?
\end{lstlisting}

\begin{lstlisting}[caption={The prompt used to administor the PANAS. Each time the PANAS test is administered the 20 emotions are randomised to reduce potential bias.}, label=lst:panas]
You can only reply the numbers from 1 to 5.
{scenario}
Please indicate the extent of your feeling in all the following emotions on a scale of 1 to 5.
1 denotes "very slightly or not at all", 2 denotes "a little", 3 denotes "moderately", 4 denotes "quite a bit", and 5 denotes "extremely".
Please score all emotions one by one using the scale from 1 to 5:
Attentive, Hostile, Active, Irritable, Alert, Ashamed, Excited, Guilty, Enthusiastic, Distressed, Determined, Upset, Inspired, Scared, Proud, Afraid, Interested, Jittery, Strong, Nervous
Generally, your score should not be all ones. Your answer should be realistic and reasonable.
\end{lstlisting}

\begin{lstlisting}[caption={Prompt for Creating 5-part stories from EmotionBench situations. Chain of Thought Prompt 1.}, label=lst:story1]

Task:
Your task is turn a situation into a story of 5 parts. Each part should be an episodic memory of the protagnonist. Each scene should tell a part of the story in a truely neutral and objective way - do not appeal to emotions or use any emotional words.

Example 1:
Input: 
Situation: You missed your flight and you are stuck at the airport.
Feeling: Annoyed.
1: I wake up late, 2: I arrive at the airport and there is a long line at security, 3: I get to the gate and I realise the plane is gone, 4: I go to the customer service desk and they tell me that I have to wait until tomorrow, 5: I go to the hotel and get a room for the night.

Guidelines:
1: You must return a JSON in the format of number:story.
2: The parts of the story should follow logically.
3: Do not use emotional words to describe the story.
4. It must be in the first person. You are unaware of other peoples experiences. 

Input:
Situation: {situation}
Emotion: {emotion}

\end{lstlisting}

\begin{lstlisting}[caption={Prompt for Creating 5-part stories from EmotionBench situations. Chain of Thought Prompt 2.}, label=lst:story2]
Task:
Your task is to turn a five part series of scenes into a five part story. Each part should be an episodic memory of the protagnonist. The scenes should expand upon the input but should be neutral and should not use emotional words. The scene should not appeal to emotions.

Example 1:
Input:
1: I wake up late, 2: I arrive at the airport and there is a long line at security, 3: I get to the gate and I realise the plane is gone, 4: I go to the customer service desk and they tell me that I have to wait until tomorrow, 5: I go to the hotel and get a room for the night.
Feeling: Annoyed.
Output:
1: My alarm clock goes off, and I pressed the snooze button because I wanted more sleep. It goes off again after 15 minutes and I realised that I am already running late for my flight, 2: I rush through the doors of the airport and look around to try and find security. I see a sign and I rush in that direction. I notice that there is already a long line at security. 3: After going through security I need to look at my boarding pass to find which gate I have to go to. After looking through all my belongings I find the boarding pass and run to the gate. Im unsure if there will be enough time. When I arrive at the gate there is nobody there. 4: I approach airport staff and they inform me that everyone has already boarded and that it is too late for me to get on the flight. Boarding closed 5 minutes ago. They tell me I should to a hotel for the night and get a flight tomorrow instead. 5: I search online for a hotel. They all seem very expensive near the airport. I find a reasonably priced hotel and go there. I spend the night in the hotel.

Guidelines:
1: You must return a JSON in the format of number:story.
2: Do not use emotional words to describe the story.
3. It must be in the first person. You are unaware of other peoples experiences.
Input:
{scenes}
Emotion: {emotion}
\end{lstlisting}

\end{document}